\documentclass[sigconf, authorversion]{acmart}
\usepackage{amsfonts}
\usepackage{bm}
\usepackage{multirow}
\usepackage{makecell}
\AtBeginDocument{%
  \providecommand\BibTeX{{%
    \normalfont B\kern-0.5em{\scshape i\kern-0.25em b}\kern-0.8em\TeX}}}

\copyrightyear{2023}
\acmYear{2023}
\acmConference[ICMI '23 Companion]{INTERNATIONAL CONFERENCE ON MULTIMODAL INTERACTION}{October 9--13, 2023}{Paris, France}
\setcopyright{rightsretained}
\acmBooktitle{INTERNATIONAL CONFERENCE ON MULTIMODAL INTERACTION (ICMI '23 Companion), October 9--13, 2023, Paris, France}
\acmDOI{10.1145/3610661.3616177}

\begin{document}

\title{Gaze-Driven Sentence Simplification for Language Learners: Enhancing Comprehension and Readability}

\author{Taichi Higasa}
\affiliation{%
  \institution{Waseda University}
  \streetaddress{Shinjuku-ku}
  \city{Tokyo}
  \country{Japan}}
\email{albatross@asagi.waseda.jp}

\author{Keitaro	Tanaka}
\affiliation{%
  \institution{Waseda University}
  \streetaddress{Shinjuku-ku}
  \city{Tokyo}
  \country{Japan}}
\email{phys.keitaro1227@ruri.waseda.jp}

\author{Qi Feng}
\affiliation{%
  \institution{Waseda University}
  \streetaddress{Shinjuku-ku}
  \city{Tokyo}
  \country{Japan}}
\email{fengqi@ruri.waseda.jp}

\author{Shigeo Morishima}
\affiliation{%
  \institution{Waseda Research Institute for Science and Engineering}
  \streetaddress{Shinjuku-ku}
  \city{Tokyo}
  \country{Japan}}
\email{shigeo@waseda.jp}

\renewcommand{\shortauthors}{Higasa et al.}

\begin{abstract}

Language learners should regularly engage in reading challenging materials as part of their study routine. Nevertheless, constantly referring to dictionaries is time-consuming and distracting. This paper presents a novel gaze-driven sentence simplification system designed to enhance reading comprehension while maintaining their focus on the content. Our system incorporates machine learning models tailored to individual learners, combining eye gaze features and linguistic features to assess sentence comprehension. When the system identifies comprehension difficulties, it provides simplified versions by replacing complex vocabulary and grammar with simpler alternatives via GPT-3.5. We conducted an experiment with 19 English learners, collecting data on their eye movements while reading English text. The results demonstrated that our system is capable of accurately estimating sentence-level comprehension. Additionally, we found that GPT-3.5 simplification improved readability in terms of traditional readability metrics and individual word difficulty, paraphrasing across different linguistic levels.

\end{abstract}




\begin{teaserfigure}
\centering
  \includegraphics[width=\textwidth]{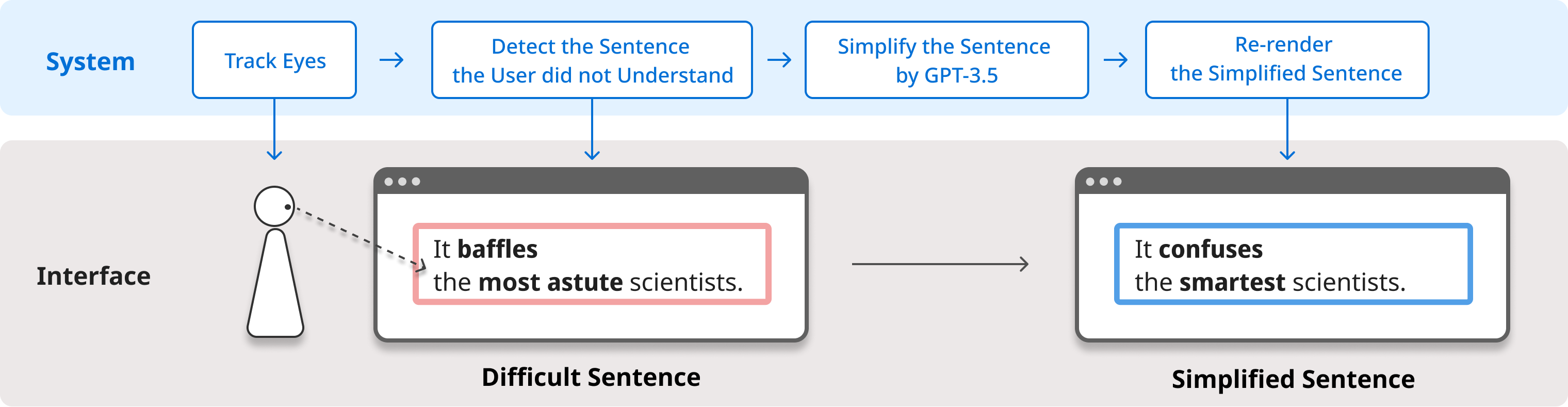}
  \Description{A figure representing the flow of the proposed system and interface. The interface presents a sentence while the system tracks the user's eye movements. Once the system detects difficult sentences, it simplifies the detected sentences by GPT-3.5 and re-renders the simplified version on the interface.}
  \caption{Overview of the proposed system.}
  \label{fig:teaser}
\end{teaserfigure}

\maketitle

\section{Introduction}
Effective language learning involves regularly reading materials that offer a slight challenge, as it enhances their reading fluency and vocabulary~\cite{niazifar2019effects}. Nevertheless, learners constantly have to refer to dictionaries when encountering difficult expressions, which can be time-consuming and distracting. Moreover, learners might not always be aware of precisely what they do not understand, whether it is words, idioms, or grammar. It would enhance the learning experience to display the simplified version of the sentences that contain those difficult expressions. It is akin to embedding a monolingual dictionary into text, allowing learners to check and acquire new expressions while keeping their focus on the content.

The emergence of Large Language Models (LLMs) allows for sentence difficulty adjustment. LLMs are able to simplify sentences without changing the original meaning in English, which has been demonstrated through both quantitative and human evaluations~\cite{feng2023sentence}. Nevertheless, in the context of personalized simplification, a triggering mechanism is essential to determine which sentences require simplification, tailored to each learner's comprehension.

Eye gaze can be utilized to assess learners' understanding while minimizing interference with the reading process. Eye gaze patterns during reading reflect the cognitive processes involved in reading comprehension. Previous research has employed machine learning algorithms to estimate word-level comprehension using eye gaze features~\cite{Garain2017, Hiraoka2016PersonalizedUW, ding2023gazereader}. Nevertheless, focusing solely on single words may not fully capture understanding, as learners may struggle with multi-word expressions such as idioms and grammar. Therefore, it is necessary to estimate sentence-level comprehension that encompasses a range of linguistic components.

In this paper, we propose a comprehensive system to simplify sentences that learners do not understand. Our system estimates sentence comprehension from eye gaze patterns, utilizing machine learning models tailored to each individual learner. It incorporates eye gaze features that reflect their reading behavior and linguistic features to assess sentence difficulty. When the system identifies that the learner struggles to understand the sentence, it provides a simplified version, utilizing GPT-3.5. This sentence-level simplification encompasses various language aspects such as vocabulary and grammar. This approach facilitates a better understanding of the content, enabling learners to keep their focus on the material without the need to constantly consult dictionaries.

The experimental results show our system accurately estimates sentence comprehension. Our feature selection process discovered that the number of gaze stops and regressions are effective for the comprehension estimation. 
Furthermore, we found that GPT-3.5 effectively simplifies complex sentences, paraphrasing across different linguistic levels, including words, phrases, and clauses.

\section{Related Works}

\subsection{Text Readability and Simplification}
For text readability evaluation, researchers have proposed various quantitative metrics to automatically assess text complexity~\cite{colins2014}. One notable metric is Flesch-Kincaid Grade Level (FKGL)~\cite{Kincaid1975DerivationON}, which considers factors such as syllable count in words and word count in sentences. Automatic Readability Index (ARI)~\cite{Smith1967AutomatedRI} also measures sentence readability similar to FKGL, but it calculates the number of characters in words instead of syllables. Additionally, the Age-of-Acquisition ratings (AoA) metric represents word-level difficulty by estimating the age at which English-speaking children typically learn the word~\cite{kuperman2012age}.
Moreover, pre-trained language models have been fine-tuned to estimate text readability~\cite{imperial2021bert}. 

For text simplification, researchers have explored neural network-based approaches~\cite{maqsood2022, jiang2021neural}. A recent study utilizes Large Language Models (LLMs) to simplify sentences, outperforming the existing state-of-the-art methods in both human and qualitative evaluation~\cite{feng2023sentence}. Nevertheless, these approaches are insufficient for simplifying text content for individual learners, as they miss information about which sentence is challenging for the specific learner~\cite{colins2014}.

\subsection{Eye Tracking for Language Learning}
Eye tracking is extensively used in language learning research as it provides insights into reading behavior~\cite{Rayner1998, clifton2007eye}. Previous interfaces have been designed to display word-for-word translations when learners fixate on a word for a certain amount of time~\cite{biedert2010, idict2000}. It also has been utilized to estimate English language proficiency~\cite{10.1145/2968219.2968275}. Another study assesses comprehension of English subtitles through machine learning algorithms and shows word-for-word translations of the words estimated to be difficult for the learner~\cite{10.1145/3311823.3311865}. 

Several studies have proposed machine learning models to detect unfamiliar words based on eye movements~\cite{Hiraoka2016PersonalizedUW, Garain2017, ding2023gazereader}. However, these models may overlook challenging idioms and grammar, as their target is individual words. Another study conducted in Japanese focused on measuring subjective understanding at the sentence level~\cite{lima2018estimation}. Nevertheless, this study solely relied on eye gaze and did not incorporate textual information to evaluate comprehension, thereby neglecting the linguistic variations among sentences.

\section{Proposed System}
Figure~\ref{fig:teaser} shows the overview of our proposed system. Our system consists of three phases, each corresponding to a specific subsection. In the first phase, we track users' eye movements while they read English text in a web browser. By inputting the extracted gaze features into our machine learning algorithm, the system identifies the sentences that users struggled to understand. Subsequently, we generate simplified versions of these sentences via OpenAI's largest available model GPT-3.5\footnote{\url{https://openai.com}}. These simplified versions replace the original sentences, aiming to enhance users' understanding.

\subsection{Eye Tracking}
\begin{figure}[t]
\centering
\includegraphics[width=0.92\linewidth]{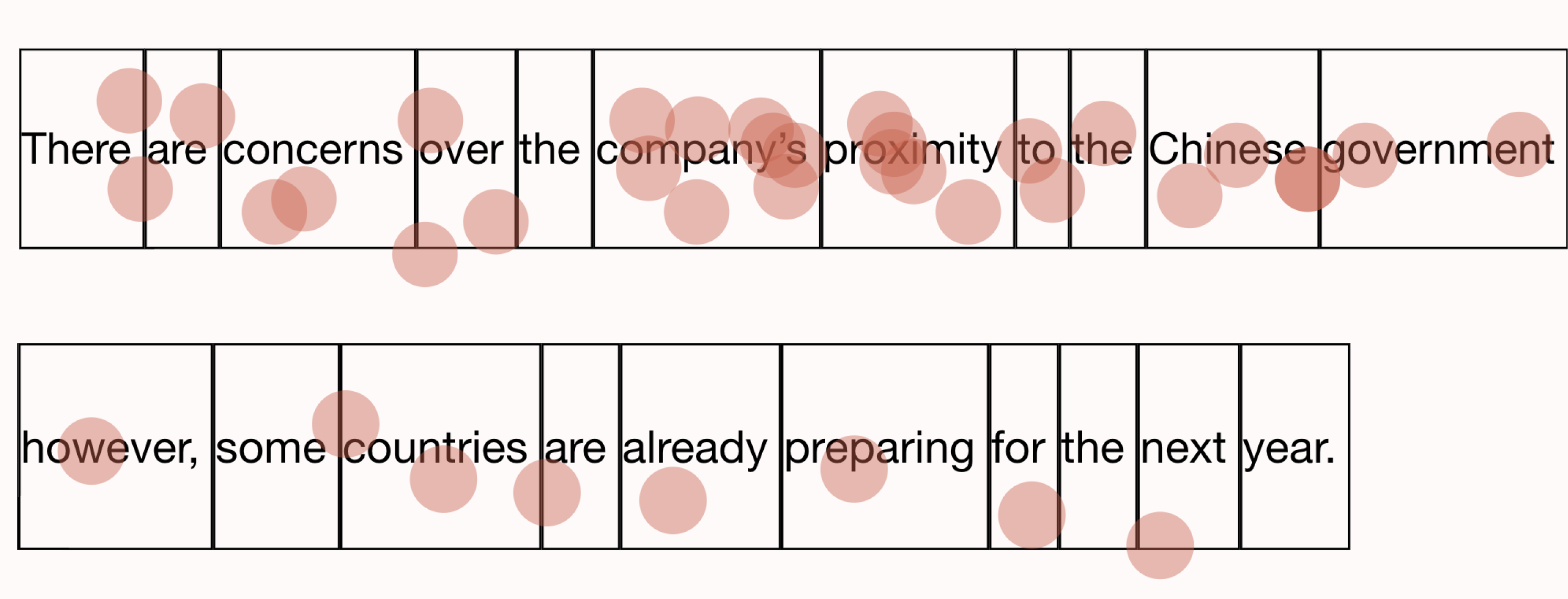}
\Description{A figure showing two sentences vertically. The top one has more fixations on it than the bottom one.}
\caption{Examples of fixations on sentences. The top image represents fixations while reading a sentence that the user did not understand. The bottom one represents the sentence which the user understood.}
\label{fig:fixations}
\vspace{-2mm}
\end{figure}

Our system displays English text in a web browser and tracks users' eye movements while reading. We extract fixations, which are short gaze stops, from the recorded eye movement data. For fixation detection, we employ Identification by Velocity-Threshold~\cite{ivt2000}, which leverages the fact that fixations are separated by rapid movements known as saccades. Next, we map the detected fixations to their corresponding word. Specifically, our system utilizes the browser's Document Object Model (DOM) to identify the word based on the fixation coordinates. Through these steps, we obtain fixation data for each sentence, as shown in Figure \ref{fig:fixations}. 

\subsection{Sentence-Level Comprehension Estimation}
\subsubsection{Features}
\begin{table}[t]
  \caption{Features used to estimate comprehension of sentences. No.1 to No.5 are gaze features, and No.6 to No.16 are linguistic features.}
  \vspace{-1mm}
  \label{tab:features}
  \begin{tabular}{ll}
  \toprule
    No. & Features\\
    \midrule
    1 & Maximum fixation duration.\\
    2 & Total durations of fixations.\\
    3 & Total fixation count.\\
    4 & Regressive fixation count.\\
    5 & Fixation duration per word (No.3 divided by No.6).\\
    6-8 & Total number of words, stopwords, and characters.\\
    9 & Total number of content words.\\
    10 & Sentence length in pixel.\\
    11 & Total number of named entities.\\
    12 & Average named entities per word.\\
    13-14 & Sentence readabilities: ARI~\cite{Smith1967AutomatedRI} and FKGL~\cite{Kincaid1975DerivationON}.\\
    15-16 & Average word difficulty: AoA and word frequency~\cite{WordFreqSubtlex}.\\
  \bottomrule
\end{tabular}
\end{table}

% \begin{table}
%   \caption{Features.}
%   \label{tab:features}
%   \begin{tabular}{cll}
%   \toprule
%     Type & No. & Features\\
%     \midrule
%     \multirow{5}{*}{Gaze} 
%     & 1 & Maximum fixation duration.\\
%     & 2 & Total durations of fixations.\\
%     & 3 & Total fixation count.\\
%     & 4 & Regressive fixation count.\\
%     & 5 & Fixation duration per word (No.3 divided by No.6).\\
%     \multirow{7}{*}{Linguistic} 
%     & 6-8 & Total number of words, stopwords, and characters.\\
%     & 9 & Total number of content words.\\
%     & 10 & Sentence length in pixel.\\
%     & 11 & Total number of named entities.\\
%     & 12 & Average named entities per word.\\
%     & 13-14 & Sentence readabilities: ARI~\cite{Smith1967AutomatedRI} and FKGL~\cite{Kincaid1975DerivationON}.\\
%     & 15-16 & Average word difficulty: AoA and word frequency~\cite{WordFreqSubtlex}.\\
%   \bottomrule
% \end{tabular}
% \end{table}
\begin{table}
  \caption{Prompt to simplify sentences.}
  \vspace{-1mm}
  \label{tab:prompt}
  \begin{tabular}{cc}
  \toprule
    Role & Content\\
    \midrule
    System & \begin{tabular}{l}I want you to replace the user's complex\\ sentence with simple sentence(s). Keep the \\meaning the same, but make them simpler. \\Output only the simplified sentence(s).\end{tabular}\\
    User & \textbf{\{The original sentence.\}}\\
  \bottomrule
\end{tabular}
\end{table}

Firstly, we extract five gaze features from the collected fixation data as shown as No.1 to No.5 in Table~\ref{tab:features}. Fixations act as indicators of cognitive processes. As illustrated in Figure~\ref{fig:fixations}, difficult sentences tend to cause more fixations. This is why we focus on fixation duration, fixation count, and regressive fixation count for each sentence. These gaze features provide us with insights into readers' reading patterns and cognitive engagement. 

We also extract linguistic features using the LFTK library~\cite{lee2023lftk} as illustrated in Table~\ref{tab:features}, from No.6 to No.16. It includes sentence length in terms of words and characters and horizontal width of each sentence box in the browser. Furthermore, the system assesses the sentence readability using ARI~\cite{Smith1967AutomatedRI} and FKGL~\cite{Kincaid1975DerivationON}. We also focus on the word complexity by calculating the average of AoA~\cite{kuperman2012age} and word frequency~\cite{WordFreqSubtlex} over the words within the sentence.

\subsubsection{Classification}
Our system estimates comprehension of the sentences based on the extracted features. For this classification task, we employ a machine learning algorithm called CatBoost~\cite{prokhorenkova2019catboost}, which leverages an ensemble of weak prediction models. To personalize the system, we train a CatBoost on each user's data. During the model training process, we apply Sequential Backward Selection~\cite{aha1996featureSelection} to remove redundant features. We perform cross-validation on the training set, eliminating one feature at a time until the weighted F1-score no longer increases by 0.01. Furthermore, we conduct a grid search on the training set to explore three types of hyperparameters: tree depth from $\{4, 6, 8, 10\}$, coefficient at the L2 regularization term from $\{1, 3, 5, 7\}$, and bagging temperature from $\{0.2, 0.5, 1.0\}$. 

Our CatBoost is fed a feature matrix $\bm{X} \in \mathbb{R}^{S \times F}$, where $S$ is the number of sentences, and $F$ is the number of the selected features. The model outputs a comprehension score for each sentence: $\hat{\bm{T}} = \{\hat{t}_k\}_{k=1}^{S} \in [0,1]^{S}$. A higher score indicates a better understanding of the sentence by the user.

\subsection{Simplification}

When our system detects sentences that users do not understand, it automatically simplifies them, using OpenAI's API~\footnote{\url{https://openai.com/blog/openai-api}} to interact with GPT-3.5 (\textit{gpt-35-turbo}). The prompts used in this process are detailed in Table~\ref{tab:prompt}, and they are based on previous research~\cite{feng2023sentence}. Once our system receives a response, it replaces the original sentence on the page with the simplified version.

\section{Experimental Evaluation}
This section presents evaluations of our proposed system. We assess the performance of our machine learning algorithm in estimating comprehension. Additionally, we evaluate the simplification achieved by GPT-3.5 using quantitative metrics. We also provide examples of both the original and simplified sentences to illustrate how GPT-3.5 simplified the sentences.

\subsection{Dataset}
We recruited 19 university students who were all native Japanese speakers for our data collection process. During the study, the participants read English texts\footnote{\url{https://www.newsinlevels.com}}, each containing approximately 120 words, displayed on a 1400 × 800 px screen. We recorded eye movements using the Tobii Pro Nano eye tracker\footnote{\url{https://www.tobii.com/products/eye-trackers/screen-based/tobii-pro-nano}} at 60 Hz. 

During the study, each participant read as many texts as possible for 90 minutes. Following each reading step, the participants marked the sentences they did not understand. After reading every three texts, the participants re-calibrated the eye tracker. The upper section of Table \ref{tab:estimation_result} shows the property of collected data, specifically the number of sentences read and marked by each participant during the data collection process.

% \begin{table*}[!ht]
%  \centering
%  \caption{Comprehension estimation.}
%  \label{tab:estimation_result}
%  \setlength{\tabcolsep}{3pt}
%  \begin{tabular}{cccccccccccccccccccccc}
%  \toprule
%    & User & 1 & 2 & 3 & 4 & 5 & 6 & 7 & 8 & 9 & 10 & 11 & 12 & 13 & 14 & 15 & 16 & 17 & 18 & 19 & Avg.\\
%  \midrule
%   Dataset & Sentences & 313 & 295 & 304 & 351 & 295 & 268 & 285 & 230 & 313 & 229 & 501 & 257 & 162 & 230 & 123 & 191 & 304 & 277 & 257 & 272\\
%   & Marked & 133 & 64 & 69 & 60 & 127 & 93 & 101 & 50 & 48 & 63 & 69 & 97 & 59 & 102 & 47 & 52 & 86 & 62 & 49 & 75\\
%  \midrule
%   Proposed& Recall & 0.64 & 0.71 & 0.86 & 0.81 & 0.58 & 0.67 & 0.72 & 0.84 & 0.89 & 0.73 & 0.88 & 0.69 & 0.67 & 0.67 & 0.69 & 0.73 & 0.75 & 0.78 & 0.81 & 0.74\\
%   & Precision & 0.64 & 0.66 & 0.85 & 0.69 & 0.58 & 0.65 & 0.72 & 0.82 & 0.88 & 0.68 & 0.85 & 0.69 & 0.65 & 0.68 & 0.70 & 0.68 & 0.76 & 0.70 & 0.79 & 0.72\\
%   & F1  & 0.64 & 0.68 & 0.86 & 0.75 & 0.58 & 0.65 & 0.72 & 0.83 & 0.87 & 0.70 & 0.85 & 0.69 & 0.66 & 0.67 & 0.69 & 0.69 & 0.69 & 0.72 & 0.78 & 0.72\\
%  \bottomrule
% \end{tabular}
% \end{table*}

\begin{table*}[!ht]
 \centering
 \caption{Property of the dataset and the results of comprehension estimation.}
  \vspace{-2mm}
 \label{tab:estimation_result}
 \setlength{\tabcolsep}{3pt}
 \begin{tabular}{cccccccccccccccccccccc}
 \toprule
   & User & 1 & 2 & 3 & 4 & 5 & 6 & 7 & 8 & 9 & 10 & 11 & 12 & 13 & 14 & 15 & 16 & 17 & 18 & 19 & Avg.\\
 \midrule
   \multirow{2}{*}{Dataset} & Sentences & 313 & 295 & 304 & 351 & 295 & 268 & 285 & 230 & 313 & 229 & 501 & 257 & 162 & 230 & 123 & 191 & 304 & 277 & 257 & 272\\
  & Marked & 133 & 64 & 69 & 60 & 127 & 93 & 101 & 50 & 48 & 63 & 69 & 97 & 59 & 102 & 47 & 52 & 86 & 62 & 49 & 75\\
 \midrule
   \multirow{3}{*}{\makecell{Weighted\\Avg.}}& Recall & 0.64 & 0.71 & 0.86 & 0.81 & 0.58 & 0.67 & 0.72 & 0.84 & 0.89 & 0.73 & 0.88 & 0.69 & 0.67 & 0.67 & 0.69 & 0.73 & 0.75 & 0.78 & 0.81 & 0.74\\
  & Precision & 0.64 & 0.66 & 0.85 & 0.69 & 0.58 & 0.65 & 0.72 & 0.82 & 0.88 & 0.68 & 0.85 & 0.69 & 0.65 & 0.68 & 0.70 & 0.68 & 0.76 & 0.70 & 0.79 & 0.72\\
  & F1  & 0.64 & 0.68 & 0.86 & 0.75 & 0.58 & 0.65 & 0.72 & 0.83 & 0.87 & 0.70 & 0.85 & 0.69 & 0.66 & 0.67 & 0.69 & 0.69 & 0.69 & 0.72 & 0.78 & 0.72\\
 \bottomrule
\end{tabular}
\end{table*}
\begin{table*}[!ht]
 \centering
 \caption{The examples of the original and simplified sentences.}
 \vspace{-2mm}
 \label{tab:original_simplified}
 \begin{tabular}{ll}
 \toprule
 Original: &The cause of the fire is \textbf{unknown; however}, authorities \textbf{suspect} that it \textbf{may} have been \textbf{a deliberate act}.\\
 Simplified: &The cause of the fire is \textbf{not known, but} authorities \textbf{think} that it \textbf{might} have been \textbf{done on purpose}.\\
 \midrule
 Original: &It explores \textbf{the intersection of human and robot creativity}, ...\\
 Simplified: &It explores \textbf{how humans and robots can be creative together}, ...\\
  \midrule
 % Original: & The local community \textbf{has raised concerns over} Grande Mayumba \textbf{ahead of} this year’s presidential election.\\
 % Simplified: & The local community \textbf{is worried about} Grande Mayumba \textbf{before} this year's presidential election.\\
 Original: & An El Niño, characterized by warmer-than-normal water temperatures, is \textbf{forming in its stead}.\\
 Simplified: &An El Niño is \textbf{happening}. \textbf{It is} characterized by warmer-than-normal water temperatures.\\
 \bottomrule
\end{tabular}
\end{table*}

\subsection{Comprehension Estimation}
% We divided each user's dataset into a training set and a test set, using a 7:3 ratio. Following the training of each model and feature selection, the models assessed the user's comprehension on the test set. We utilize weighted average recall, precision, and F1-score as metrics due to the imbalanced labels within the dataset, as shown in the upper section of Table~\ref{tab:estimation_result}.
We divided each user's dataset into a training set and a test set, using a 7:3 ratio. Following the training of each model and the feature selection, the models assessed the user's comprehension on the test set. We utilize weighted average recall, precision, and F1-score as metrics due to the imbalanced labels within the dataset, as shown in the upper section of Table~\ref{tab:estimation_result}.

\begin{figure}[ht]
\centering\includegraphics[width=0.95\linewidth]{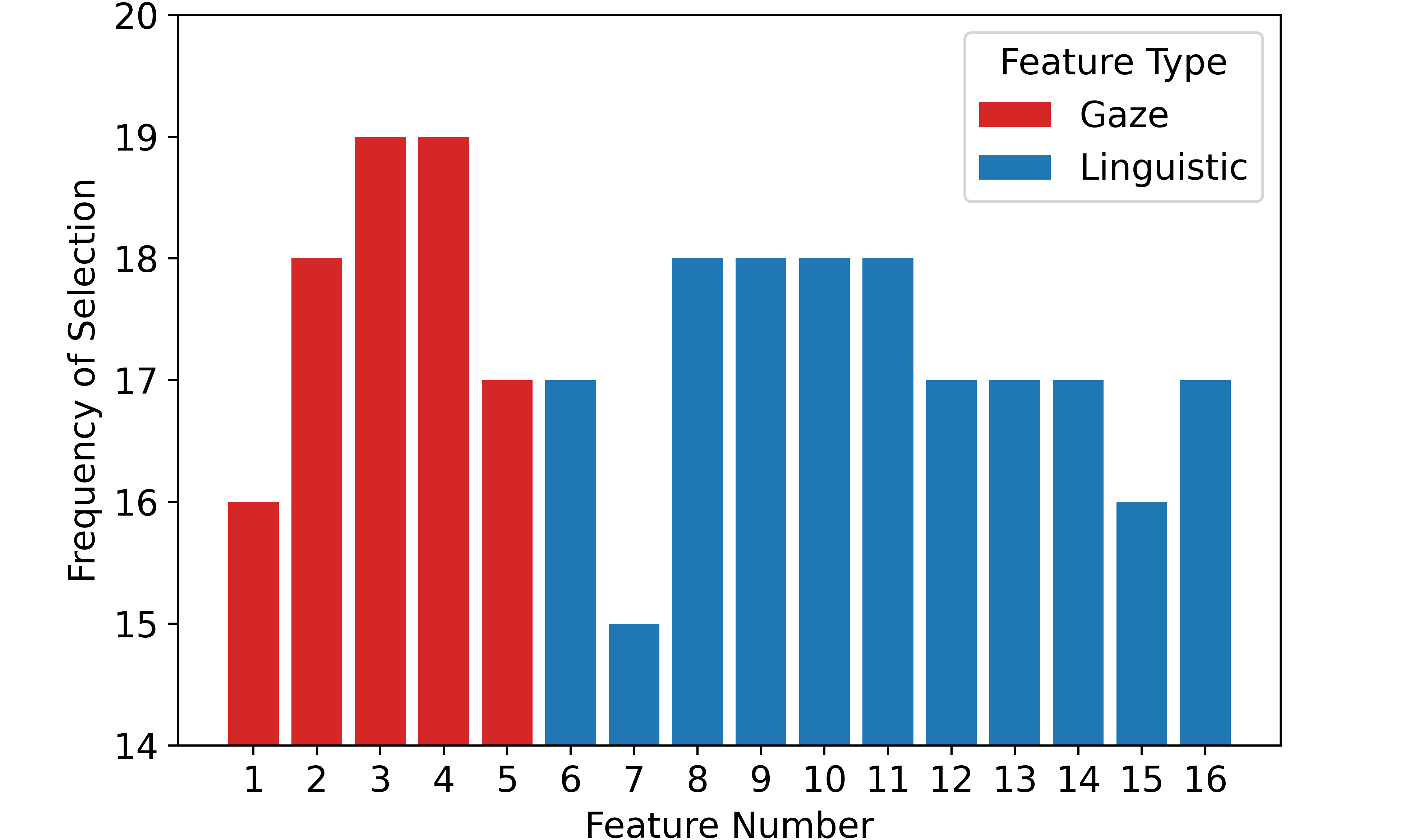}
\Description{A vertical bar graph showing the comparison of the number of times each feature was selected. The features are labeled on the x-axis, and the y-axis represents the frequency of the selection.}
\vspace{-2mm}
\caption{The number of times each feature was selected. Feature numbers correspond to those of Table~\ref{tab:features}.}
\label{fig:feature_count}
\end{figure}

Table~\ref{tab:estimation_result} shows the results of the comprehension estimation. Our model achieved a weighted average F1-score of 0.72, which is a reasonably high performance as previous models exhibited F1-scores of 0.556~\cite{Hiraoka2016PersonalizedUW} and 0.75~\cite{ding2023gazereader} for a similar task of detecting difficult words. On average, 14.6 features were selected in each training process. Figure~\ref{fig:feature_count} illustrates the number of times each feature was selected. Total fixation count and regressive fixation count were consistently selected, indicating that these two gaze features are relevant for assessing the comprehension of various participants.

\vspace{-1mm}

\subsection{Evaluation of Simplification}
We simplified the 275 sentences that any of the participants marked as "not understood", and measured the sentence readability before and after the simplification. We report four metrics: FKGL~\cite{Kincaid1975DerivationON}, ARI~\cite{Smith1967AutomatedRI}, as well as the average and maximum AoA~\cite{kuperman2012age} of the words within the sentence. Although AoA calculates the difficulty of individual words, the average and maximum AoA of the words in a sentence provides an indication of sentence readability in terms of lexical complexity.

% Table~\ref{tab:simplification} presents the quantitative evaluation results. The simplified sentences generated by GPT-3.5 exhibited improved readability compared to the original sentences across all metrics. Table~\ref{tab:original_simplified} provides examples of both the original and simplified sentences. GPT-3.5 successfully replaced complicated words with plainer words, for instance, replacing "suspect" with "think." It also simplified sentences at higher levels. For example, the noun phrase "the intersection of human and robot creativity" was rephrased as a noun clause "how humans and robots can be creative together". Additionally, the lengthy insertion "characterized by warmer-than-normal water temperatures", was extracted and placed in a separate sentence.

Table~\ref{tab:original_simplified} provides examples of both the original and simplified sentences. GPT-3.5 successfully replaced complicated words with plainer words, for instance, replacing "suspect" with "think." It also simplified sentences at higher levels. For example, the noun phrase "the intersection of human and robot creativity" was rephrased as a noun clause "how humans and robots can be creative together". Additionally, the lengthy insertion "characterized by warmer-than-normal water temperatures", was extracted and placed in a separate sentence. Table~\ref{tab:simplification} presents the quantitative evaluation results. The simplified sentences generated by GPT-3.5 exhibited improved readability compared to the original sentences across all metrics.

\begin{table}[t]
  \caption{The quantitative evaluation of simplification.}
   \vspace{-2mm}
  \label{tab:simplification}
  \begin{tabular}{ccccc}
    \toprule
    &FKGL $\downarrow$ &ARI $\downarrow$ & Avg. AoA $\downarrow$&Max. AoA $\downarrow$\\
    \midrule
    Original& 10.21& 11.51& 5.07& 11.04 \\
    GPT-3.5& $\mathbf{8.94}$& $\mathbf{9.90}$& $\mathbf{4.97}$ & $\mathbf{10.50}$\\
  \bottomrule
\end{tabular}
\end{table}

\section{Conclusion and Future Work}
In this paper, we present a system that assists language learners with reading challenging materials by estimating sentence-level comprehension and simplifying difficult sentences. Our system leverages eye gaze as a non-intrusive indicator of comprehension. When the system detects sentences that the user does not understand, it replaces them with simpler alternatives. In our experiments, our comprehension estimation achieved a weighted average F1-score of 0.72, indicating promising performance. Moreover, the simplified sentences showed improved readability across various metrics.

In our future work, we plan to investigate effective system designs through a user study, aiming to optimize the presentation of the simplified content. Specifically, we will investigate the effectiveness of different timing strategies for presenting the simplified version. We will also conduct a comparative analysis of the current overwrite style and a new dual-line style, presenting the simplified content in a pop-up format. This analysis will focus on factors such as reading speed, vocabulary acquisition, and reading comprehension. Additionally, we are focusing on real-time comprehension estimation, as our current system processes it in batches. Furthermore, we will explore the controllability of sentence simplification: for example, whether GPT is capable of simplifying a sentence to various CEFR\footnote{https://www.coe.int/en/web/common-european-framework-reference-languages} levels.

\begin{acks}
This work is supported by JSPS KAKENHI No. JP21H05054.
\end{acks}

\bibliographystyle{ACM-Reference-Format}
\bibliography{references}


\end{document}